\documentclass[conference]{IEEEtran}
\def\IEEEtitletopspace{18pt}
\IEEEoverridecommandlockouts
\usepackage{multirow}
\usepackage{caption}
\usepackage{cite}
\usepackage{amsmath,amssymb,amsfonts}
\usepackage{algorithmic}
\usepackage{algorithm}
\usepackage{graphicx}
\usepackage{textcomp}
\usepackage{xcolor}
\def\BibTeX{{\rm B\kern-.05em{\sc i\kern-.025em b}\kern-.08em
    T\kern-.1667em\lower.7ex\hbox{E}\kern-.125emX}}

\usepackage{cuted}
\usepackage{subcaption}
\usepackage{array}
\usepackage{comment}
\usepackage{balance}

\newcolumntype{C}[1]{>{\centering\arraybackslash}m{#1}}
\renewcommand\IEEEkeywordsname{Keywords}
    
\begin{document}
\title{Multimodal RGB–Infrared Combination for UAV-Based Wildfire Segmentation: A Comparative Study on FLAME3
\thanks{This research is sponsored by national funds through Fundação para a Ciência e a Tecnologia (FCT), under projects UID/00048/2025 (DOI: 10.54499/UIDB/00048/2025), UID/00285/2025 (DOI: 10.54499/UID/00285/2025) and LA/P/0112/2020 (DOI: 10.54499/LA/P/0112/2020), and co-financed by the European Regional Development Fund through Innovation and Digital Transition Thematic Programme (COMPETE 2030) of Portugal 2030, Project FireSys (COMPETE2030-FEDER-02230700).}
}

\author{
\IEEEauthorblockN{
Matheus F. Kovaleski\IEEEauthorrefmark{1},
Luís Garrote\IEEEauthorrefmark{1},
Cristiano Premebida\IEEEauthorrefmark{1},
Jérôme Mendes\IEEEauthorrefmark{2,3}, and
João Ruivo Paulo\IEEEauthorrefmark{1}
}

\IEEEauthorblockA{\IEEEauthorrefmark{1}
Institute of Systems and Robotics, Department of Electrical and Computer Engineering, University of Coimbra, Portugal\\
Email: \{matheus.kovaleski, garrote, cpremebida, jpaulo\}@isr.uc.pt
}

\IEEEauthorblockA{\IEEEauthorrefmark{2}
University of Coimbra, CEMMPRE, ARISE,
Department of Mechanical Engineering, Polo II, PT-3030-788 Coimbra, Portugal\\
Email: jerome.mendes@uc.pt
}

}

\maketitle

\begin{abstract}
Unmanned Aerial Vehicles (UAVs) have emerged as a promising platform for firefighting operations due to their flexibility, low operational cost, and ability to acquire high-resolution imagery in locations that may be difficult or dangerous to access using conventional methods. Recent advances in deep learning have significantly improved the capabilities of UAV-based wildfire monitoring systems. The present work investigates RGB-infrared fusion for binary wildfire segmentation on the FLAME3 dataset. In this Study, RGB and Infrared baselines are compared with three representative fusion strategies across three segmentation architectures, including U-Net, DeepLabV3+, and SegFormer. The key motivation of this work is to analyze the contribution of each modality, evaluate the impact of fusion timing, and examine how different network architectures exploit multimodal information for UAV wildfire delineation. The findings indicate that thermal information plays a dominant role in UAV segmentation and that feature-level multimodal fusion combined with transformer-based architectures offers the most promising direction for future research.

\end{abstract}


\section{Introduction}
\label{sec.Introduction}

Wildfires have become increasingly frequent and severe in many regions of the world, causing substantial environmental, economic, and societal impacts. Climate change, prolonged droughts, and increasing human activity have contributed to the growing intensity and frequency of wildfire events, creating an urgent need for effective monitoring and rapid response systems \cite{Ghali2023Review}. Early identification of active fire fronts is critical for supporting firefighting operations, reducing burned areas, and minimizing risks to human life and infrastructure.

Unmanned Aerial Vehicles (UAVs) have emerged as a promising platform for wildfire monitoring due to their flexibility, low operational cost, and ability to acquire high-resolution imagery in locations that may be difficult or dangerous to access using conventional methods \cite{Chen2022RGBIR,Boroujeni2024Survey}. Compared with satellite-based systems, UAVs can provide imagery with higher spatial resolution and lower latency, enabling detailed observation of fire behavior and supporting near-real-time decision making during emergency operations.

Recent advances in deep learning have significantly improved the capabilities of UAV-based wildfire monitoring systems. Convolutional Neural Networks (CNNs) and Transformer-based architectures have demonstrated remarkable performance in image understanding tasks, including wildfire classification, detection, and semantic segmentation \cite{Ghali2023Review}. Among these approaches, semantic segmentation is particularly attractive because it provides pixel-level delineation of fire regions, offering more detailed spatial information than image-level classification or bounding-box detection. Architectures such as U-Net, DeepLabV3+, and Vision Transformers have achieved state-of-the-art performance in aerial wildfire segmentation tasks \cite{Morandini2021Segmentation,Ghali2022Transformer,Ghali2021VisionTransformer}.

Most existing wildfire segmentation methods rely primarily on RGB imagery. Although visible-spectrum images provide rich contextual information and detailed spatial structures, their performance may degrade under adverse conditions such as dense smoke, illumination variations, shadows, and low-contrast fire boundaries \cite{Ghali2024RGBThermal}. Thermal infrared (IR) imagery offers complementary information by directly capturing heat emissions from active fire regions and remaining relatively robust to changes in lighting conditions. Recent studies have shown that combining visible and thermal modalities can significantly improve wildfire recognition and segmentation performance by exploiting complementary visual and thermal cues \cite{Rui2023RGBT,Yue2025Multimodal}.

The recent introduction of the FLAME~3 dataset \cite{w0mz-aq48-24} represents an important step toward multimodal wildfire analysis. Unlike previous versions of the FLAME benchmark, FLAME~3 provides synchronized visible-spectrum and radiometric thermal imagery acquired from UAV platforms, enabling the investigation of multimodal learning strategies for wildfire management tasks \cite{Hopkins2024FLAME3}. The availability of radiometric thermal measurements creates new opportunities for studying how thermal information can complement conventional RGB observations in segmentation applications.

Despite the increasing interest in multimodal wildfire sensing, there is still a limited understanding of how different fusion strategies influence segmentation performance. Existing works often evaluate a single fusion approach or focus on a specific architecture, making it difficult to isolate the effects of modality selection and fusion timing. Furthermore, systematic comparisons between RGB-only, infrared-only, and multimodal approaches under a unified experimental protocol remain scarce.

To address this gap, this paper investigates RGB--infrared fusion for binary wildfire segmentation using the FLAME~3 dataset. We compare RGB-only and infrared-only baselines with three representative fusion strategies---early fusion, middle fusion, and late fusion---across different segmentation architectures, including U-Net, DeepLabV3, and SegFormer. The objective is to analyze the contribution of each modality, evaluate the impact of fusion timing, and examine how different network architectures exploit multimodal information for UAV-based wildfire segmentation.

The main contributions of this comparative study are summarized as follows:

\begin{itemize}
    \item A systematic comparison between RGB-only and infrared-only modalities for UAV wildfire segmentation using FLAME~3.
    
    \item An evaluation of early, middle, and late fusion strategies under a common experimental protocol.
    
    \item A comparative analysis of convolutional and Transformer-based segmentation architectures, namely U-Net, DeepLabV3, and SegFormer.
    
    \item An investigation of the relationship between fusion strategy, modality selection, and segmentation performance in radiometric UAV wildfire imagery.
\end{itemize}

The remainder of this paper is organized as follows. Section~\ref{sec:relatedwork} reviews related work on UAV wildfire segmentation and multimodal fusion methods. Section~\ref{sec:methodology} describes the dataset, fusion strategies, network architectures, and experimental setup. Section~\ref{sec:results} presents the experimental results and comparative evaluation. Section~\ref{sec:conclusion} concludes the paper and outlines future research directions.


\section{Related Work}
\label{sec:relatedwork}

\subsection{UAV-Based Wildfire Segmentation}

The increasing availability of UAV platforms has significantly improved the acquisition of high-resolution imagery for wildfire monitoring and emergency response. Compared with satellite-based systems, UAVs provide finer spatial resolution and greater operational flexibility, enabling detailed observation of fire fronts, smoke plumes, and burned areas in near real time \cite{Boroujeni2024Survey,Keerthinathan2023UAS}.

The development of publicly available datasets has played a fundamental role in advancing deep learning methods for wildfire analysis. Among these datasets, the FLAME benchmark introduced aerial RGB and thermal imagery with pixel-level annotations, becoming one of the most widely used datasets for wildfire detection and segmentation research \cite{Shamsoshoara2021FLAME}. More recently, FLAME~3 extended this benchmark by providing synchronized RGB and radiometric thermal imagery, allowing the investigation of multimodal learning approaches based on calibrated thermal measurements \cite{Hopkins2024FLAME3}. Additional datasets, such as FireMan-UAV-RGBT and the Boreal Forest Fire dataset, have further contributed to the study of multimodal wildfire perception in UAV environments \cite{Kularatne2024FireMan,Pesonen2025Boreal}.

Several studies have demonstrated the effectiveness of deep learning for wildfire segmentation using aerial imagery. Comprehensive reviews indicate that segmentation models consistently outperform traditional computer vision approaches by learning robust visual representations directly from image data \cite{Ghali2023Review}, \cite{Saleh2023Review}. However, accurately segmenting wildfire regions remains challenging due to small flame areas, smoke occlusions, illumination changes, and highly heterogeneous backgrounds.

\subsection{Deep Learning Architectures for Wildfire Segmentation}

Encoder--decoder convolutional architectures remain among the most widely adopted approaches for wildfire segmentation. U-Net has become a standard baseline because of its relatively low computational cost and its ability to preserve spatial information through skip connections. Previous studies demonstrated competitive segmentation performance using U-Net across different wildfire datasets and imaging conditions \cite{Morandini2021Segmentation}.

DeepLabV3+ introduced atrous convolutions and Atrous Spatial Pyramid Pooling (ASPP), enabling the extraction of multi-scale contextual information while preserving spatial resolution. This architecture has consistently achieved superior segmentation accuracy compared with conventional encoder--decoder networks in aerial wildfire imagery \cite{Morandini2021Segmentation}. More recent approaches have explored transformer-based models capable of capturing long-range dependencies through self-attention mechanisms. \cite{Ghali2021VisionTransformer,Ghali2022Transformer} demonstrated that vision transformers such as TransUNet can outperform conventional CNN architectures in wildfire segmentation tasks by jointly modeling local details and global contextual information.

Recent hybrid CNN--Transformer architectures continue to improve segmentation quality in challenging wildfire scenarios. For example, FireFormer combines convolutional feature extraction with transformer-based contextual modeling and achieved competitive performance on the FLAME dataset \cite{Tong2024FireFormer}. These developments suggest that different architectures may exploit multimodal information in distinct ways, motivating comparative evaluations across CNN-based and transformer-based segmentation models.

\subsection{RGB--Infrared Fusion for Wildfire Analysis}

While most wildfire segmentation systems rely on RGB imagery, thermal infrared sensing provides complementary information related to heat emission and active combustion. RGB images offer rich texture, shape, and contextual cues, whereas infrared imagery can highlight active fire regions even under adverse illumination conditions or partial smoke occlusion \cite{Ghali2024RGBThermal}.

Some recent studies have investigated multimodal wildfire analysis using RGB and thermal information. \cite{Morandini2021Segmentation} evaluated different image modalities for wildfire segmentation and reported that infrared and fused imagery can improve segmentation performance compared with visible imagery alone. \cite{Rui2023RGBT} proposed an adaptive RGB-T modality learning framework that jointly exploits modality-specific and shared representations, achieving substantial improvements in segmentation accuracy under day and night conditions. Similarly, \cite{Yue2025Multimodal} introduced a multimodal segmentation framework based on multilevel fusion mechanisms and demonstrated that combining RGB and thermal features improves both IoU and F1-score compared with unimodal baselines.

Despite these promising results, the literature still provides limited understanding of how different fusion strategies influence wildfire segmentation performance. Existing studies typically focus on a single architecture or a specific fusion mechanism, making it difficult to isolate the effects of modality selection and fusion timing. Furthermore, few works systematically compare RGB-only, infrared-only, and multiple fusion strategies under a unified experimental protocol. The present work addresses this gap through a comprehensive evaluation of early, middle, and late fusion approaches across U-Net, DeepLabV3, and SegFormer architectures using the FLAME~3 dataset.

\section{Methodology}
\label{sec:methodology}

This section describes the dataset, ground-truth generation procedure, multimodal fusion strategies, evaluated segmentation models, and experimental setup adopted in this study.

\subsection{Dataset}

The experiments are conducted using the FLAME~3 dataset \cite{Hopkins2024FLAME3}, a UAV-based wildfire benchmark that provides synchronized RGB imagery, thermal imagery, and radiometric temperature measurements. The dataset contains samples acquired under different wildfire conditions and was specifically designed to support multimodal wildfire analysis using visible and thermal sensing modalities.

For each sample, three data sources are available:

\begin{itemize}
    \item RGB images captured by the visible-spectrum camera;
    \item Thermal infrared images stored as colorized thermal JPEG files;
    \item Radiometric temperature maps provided as Celsius TIFF images.
\end{itemize}

Only samples containing valid RGB, thermal JPEG, and radiometric TIFF files were included in the experiments. All images were resized to $512 \times 512$ pixels before training and evaluation. RGB images were normalized using ImageNet statistics, while thermal images were independently normalized to preserve their temperature-related information.

Dataset partitioning followed a random split strategy with 70\% of the samples used for training, 15\% for validation, and 15\% for testing. The same partition was maintained across all experiments to ensure a fair comparison between architectures and fusion strategies.

\subsection{Mask Generation}

Unlike conventional segmentation datasets that provide manually annotated masks, FLAME~3 includes radiometric temperature measurements that can be used to derive active-fire regions.

Binary fire masks were generated directly from the radiometric Celsius TIFF images through temperature thresholding. Let $T(x,y)$ denote the temperature value at pixel $(x,y)$. A pixel was labeled as fire when
\begin{equation}
M(x,y)=
\begin{cases}
1, & T(x,y)\geq 80^\circ C \\
0, & \text{otherwise}
\end{cases}
\end{equation}
where $M$ represents the resulting binary segmentation mask. The threshold of $80^{\circ}$C was adopted based on the temperature-based fire/no-fire screening criterion reported for the FLAME~3 radiometric thermal data~\cite{Hopkins2024FLAME3}.

This procedure produces objective fire-region annotations directly from measured thermal information and avoids the variability associated with manual labeling. The generated masks were used as ground truth for all segmentation experiments.

\subsection{Fusion Strategies}

To investigate the contribution of visible and thermal information, five input configurations are evaluated.

\begin{itemize}
    \item \textbf{RGB}: only RGB imagery is provided to the network.
    
    \item \textbf{IR}: only thermal imagery is used as input.
    
    \item \textbf{Early Fusion}: RGB and thermal images are concatenated at the input level, producing a six-channel tensor composed of three RGB channels and three thermal channels.
    
    \item \textbf{Middle Fusion}: RGB and thermal modalities are processed through independent feature extractors. Intermediate feature maps are fused before the decoder stage, allowing modality-specific representations to be learned before integration.
    
    \item \textbf{Late Fusion}: RGB and thermal data are processed by independent segmentation networks. The final segmentation logits generated by each branch are fused through a learnable fusion layer to obtain the final prediction.
\end{itemize}

\begin{figure*}[t]
\centering
    \includegraphics[trim={1cm 0.6cm 1cm 1cm}, width=0.95\textwidth]{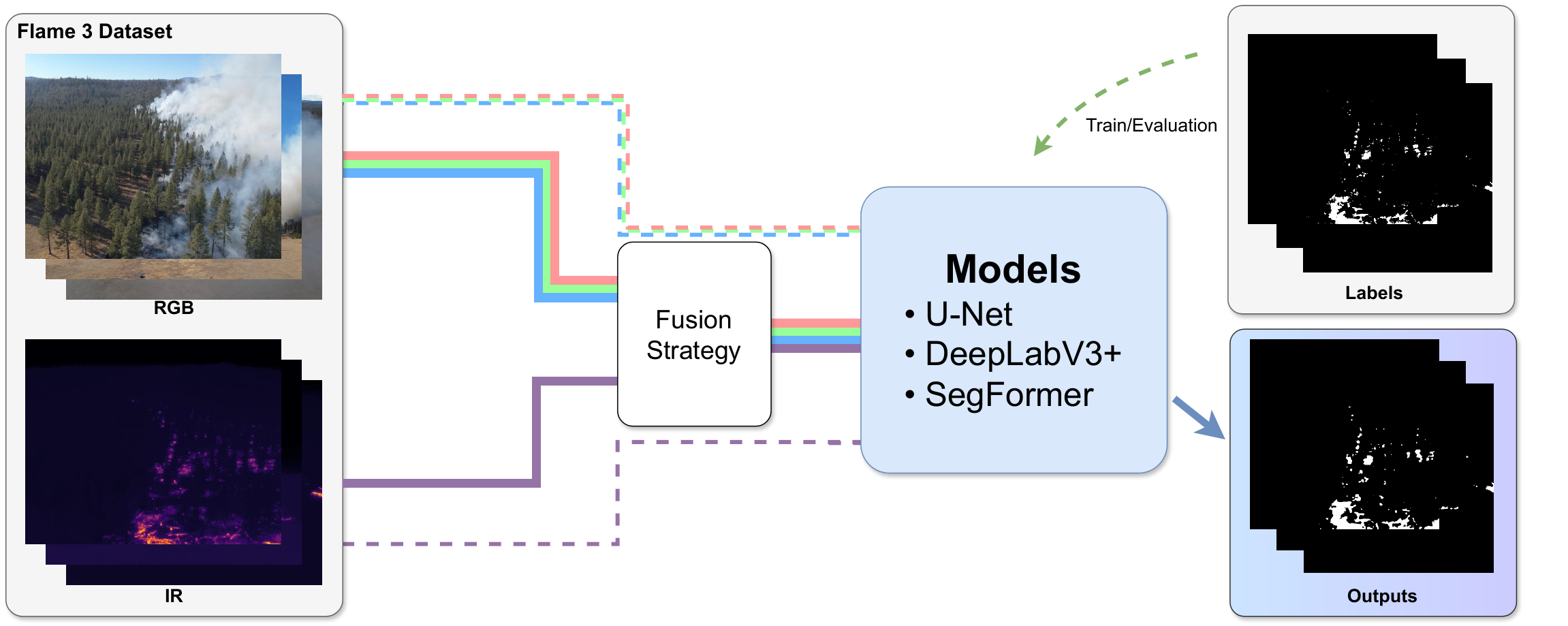}
    \caption{Overview of the proposed RGB--IR wildfire segmentation pipeline. RGB and thermal infrared UAV imagery are combined using different fusion strategies and evaluated using U-Net, DeepLabV3+, and SegFormer architectures. The generated predictions are assessed using standard semantic segmentation metrics.}
\label{fig:fusion_pipeline}
\end{figure*}

Figure~\ref{fig:fusion_pipeline} illustrates the evaluated fusion strategies. The RGB and IR configurations serve as unimodal baselines, while the early, middle, and late fusion approaches enable the analysis of how fusion timing influences segmentation performance.

\subsection{Segmentation Models}

Three semantic segmentation architectures were evaluated in this work: U-Net, DeepLabV3, and SegFormer. These models were selected due to their complementary characteristics regarding spatial-detail preservation, contextual aggregation, and global feature representation.

\subsubsection{U-Net}

U-Net \cite{Ronneberger2015UNet} is an encoder--decoder convolutional architecture that uses skip connections to preserve spatial information throughout the reconstruction process. Due to its strong localization capability, U-Net has been widely adopted in segmentation tasks involving small and sparse targets.

\subsubsection{DeepLabV3}

DeepLabV3 \cite{chen2018encoder} incorporates atrous convolutions and Atrous Spatial Pyramid Pooling (ASPP) to capture multi-scale contextual information. In this work, the MobileNet backbone version was adopted due to its favorable balance between computational efficiency and segmentation performance.

\subsubsection{SegFormer}

SegFormer \cite{Xie2021SegFormer} is a transformer-based segmentation architecture that combines hierarchical self-attention representations with a lightweight decoder. The model is capable of capturing long-range dependencies while preserving multi-scale feature representations.

By comparing these architectures, we analyze how convolutional and transformer-based models exploit visible, thermal, and fused multimodal information for wildfire segmentation.

\subsection{Training and Evaluation Protocol}
\label{sec:training_protocol}

All experiments were performed using the same training protocol to ensure a fair comparison across architectures and fusion strategies.

The models were trained using binary cross-entropy with logits loss (BCEWithLogitsLoss) and optimized with the AdamW optimizer. The learning rate was fixed at $10^{-4}$, the batch size was set to 2, and training was performed for 30 epochs. Model selection was based on the highest validation Intersection over Union (IoU).

Performance was evaluated using four standard segmentation metrics:

\begin{itemize}
    \item Precision;
    \item Recall;
    \item F1-score;
    \item Intersection over Union (IoU).
\end{itemize}

During evaluation, segmentation probabilities were converted into binary predictions using a threshold of $0.5$. All reported results correspond to the performance obtained on the held-out test set.

\section{Experimental Results}
\label{sec:results}

This section presents the experimental evaluation of RGB--IR fusion strategies for wildfire segmentation on the FLAME~3 dataset. The goal of these experiments is to analyze how visible and infrared modalities contribute to fire segmentation performance, as well as how different fusion strategies affect the behavior of multiple segmentation architectures.

We compare single-modality baselines using RGB-only and IR-only inputs with three multimodal fusion strategies: early fusion, middle fusion, and late fusion. This comparison allows us to assess whether combining RGB and infrared information improves segmentation performance over individual modalities, and whether the fusion stage influences precision, recall, F1-score, and IoU.

\subsection{Training Setup}

All models were trained for binary semantic segmentation using the common training protocol described in Section~\ref{sec:training_protocol}. The same dataset partitions, preprocessing pipeline, and optimization settings were maintained across all architectures and fusion configurations to ensure a consistent comparison.

\subsection{Evaluation Protocol}

All models were evaluated on the FLAME~3 test set using Precision, Recall, F1-score, and Intersection over Union (IoU), as described in Section~\ref{sec:training_protocol}. The comparison focuses on the contribution of each modality and on whether RGB--IR fusion improves performance over the single-modality baselines across different architectures.

\subsection{Results}

Table~\ref{tab:results} reports the wildfire segmentation results obtained on the FLAME~3 dataset.

\begin{table}[t]
\centering
\caption{Wildfire segmentation performance on the FLAME~3 test set for RGB-only, IR-only, and RGB--IR fusion configurations across the evaluated architectures. Bold values indicate the best result for each metric.}
\label{tab:results}
\setlength{\tabcolsep}{7pt}
\renewcommand{\arraystretch}{1.12}
\begin{tabular}{ll|cccc}
\noalign{\hrule height 1pt} \hline
Architecture & Modality & Prec. & Rec. & IoU & F1 \\
\hline
\hline

\multirow{5}{*}{U-Net}
& RGB           & 0.70 & 0.09 & 0.08 & 0.17  \\
& IR            & 0.74 & 0.82 & 0.62 & 0.75  \\
& RGB--IR Early & \textbf{0.92} & 0.62 & 0.58 & 0.73  \\
& RGB--IR Middle& 0.83 & 0.72 & 0.60 & 0.74  \\
& RGB--IR Late  & 0.78 & 0.78 & 0.61 & 0.74  \\
\hline

\multirow{5}{*}{DeepLabV3+}
& RGB           & 0.61 & 0.09 & 0.07 & 0.19  \\
& IR            & 0.83 & 0.64 & 0.55 & 0.69  \\
& RGB--IR Early & 0.79 & 0.68 & 0.56 & 0.71  \\
& RGB--IR Middle& 0.69 & 0.51 & 0.37 & 0.51  \\
& RGB--IR Late  & 0.87 & 0.64 & 0.58 & 0.74  \\
\hline

\multirow{5}{*}{SegFormer}
& RGB           & 0.59 & 0.19 & 0.14 & 0.24  \\
& IR            & 0.89 & 0.76 & 0.69 & 0.81  \\
& RGB--IR Early & 0.85 & 0.75 & 0.65 & 0.76  \\
& RGB--IR Middle& 0.86 & 0.84 & \textbf{0.74} & \textbf{0.84}  \\
& RGB--IR Late  & 0.83 & \textbf{0.87} & 0.74 & 0.84  \\
\noalign{\hrule height 1pt} \hline

\end{tabular}
\end{table}

\subsubsection{U-Net Results}

U-Net shows a clear performance gap between RGB-only and IR-based configurations. The RGB-only model obtains low recall and IoU, indicating that visible imagery alone is insufficient for accurately detecting fire regions in this dataset. Although RGB achieves moderate precision, the low recall shows that most fire pixels are missed.

The IR-only configuration substantially improves performance, reaching an F1-score of 0.75 and an IoU of 0.62. This indicates that the infrared modality provides more discriminative information for wildfire segmentation than RGB alone. Among the RGB--IR fusion strategies, early, middle, and late fusion achieve similar F1-scores, ranging from 0.73 to 0.74. However, none of the fusion strategies clearly surpass the IR-only baseline.

Early fusion produces the highest precision among the U-Net configurations, but this comes at the cost of lower recall. In contrast, late fusion provides a more balanced precision-recall trade-off, with recall close to the IR-only setup. Overall, these results suggest that U-Net benefits strongly from infrared information, but RGB does not provide a consistent additional gain when fused with IR.

\subsubsection{DeepLabV3+ Results}

DeepLabV3+ also performs poorly with RGB-only input, achieving low recall and IoU. Similar to U-Net, this confirms that RGB information alone is not sufficient for reliable wildfire segmentation on FLAME~3.

The IR-only configuration provides a strong improvement, with an F1-score of 0.69 and an IoU of 0.55. RGB--IR early fusion slightly improves F1-score and IoU compared with IR-only, suggesting that combining RGB and IR at the input level can provide complementary information for this architecture. However, middle fusion performs substantially worse, with an F1-score of 0.51 and an IoU of 0.37. This indicates that the intermediate fusion design is less effective for DeepLabV3+ in this setting.

The best DeepLabV3+ result is obtained with late fusion, which reaches an F1-score of 0.74 and an IoU of 0.58. This suggests that processing RGB and IR streams more independently before fusion is more suitable for DeepLabV3+ than combining intermediate features too early. Overall, DeepLabV3+ benefits from RGB--IR fusion, but its performance depends strongly on the fusion stage.

\subsubsection{SegFormer Results}

SegFormer achieves the strongest overall results among the evaluated architectures. As observed for U-Net and DeepLabV3+, the RGB-only configuration performs poorly compared with IR-based configurations. However, SegFormer obtains the best RGB-only result among the three models, with an F1-score of 0.24 and an IoU of 0.14.

The IR-only configuration provides a substantial improvement, reaching an F1-score of 0.81 and an IoU of 0.69. This confirms the strong relevance of infrared information for wildfire segmentation. RGB--IR early fusion does not improve over IR-only, suggesting that simple input-level concatenation may introduce redundant or less informative visible features.

The best SegFormer results are obtained with middle and late fusion. Middle fusion achieves the highest IoU and F1-score, while late fusion obtains the highest recall. This indicates that SegFormer is better able to exploit complementary RGB and IR information when modalities are fused after modality-specific feature extraction. The strong recall obtained by late fusion suggests that this strategy is particularly effective at detecting fire pixels, while middle fusion provides a slightly better overall balance between precision and recall.

\subsubsection{Cross-Model Analysis}

The results reveal three main findings. First, RGB-only segmentation performs poorly across all architectures. Although precision remains moderate, recall is consistently low, indicating that models trained only with visible imagery miss most fire pixels. This behavior suggests that RGB information alone is not sufficiently reliable for active wildfire segmentation in the evaluated setting.

Second, infrared information is the dominant modality for wildfire segmentation on FLAME~3. IR-only models substantially outperform RGB-only models across all architectures, with large gains in recall, F1-score, and IoU. This confirms that the infrared modality captures fire-related information that is not easily detected from visible imagery alone.

Third, the effectiveness of RGB--IR fusion depends on the architecture and fusion strategy. For U-Net, fusion does not clearly improve over the IR-only baseline, suggesting that the RGB modality provides limited additional benefit for this encoder-decoder architecture. For DeepLabV3+, late fusion provides the best result, while middle fusion performs poorly. For SegFormer, middle and late fusion achieve the strongest overall performance, indicating that transformer-based architectures may benefit more from modality-specific feature extraction before fusion.

Overall, SegFormer with RGB--IR middle fusion provides the best trade-off between precision, recall, F1-score, and IoU. However, the strong performance of IR-only models across all architectures indicates that infrared imagery is the key modality for wildfire segmentation, while RGB information should be incorporated carefully through an appropriate fusion strategy.

\section{Conclusion}
\label{sec:conclusion}

This work investigated the use of RGB-infrared fusion strategies for UAV-based wildfire segmentation using the FLAME~3 dataset. A systematic comparison has been conducted between unimodal baselines and multimodal fusion approaches across three segmentation architectures: U-Net, DeepLabV3+, and SegFormer.

The experimental results demonstrated that thermal infrared imagery provides substantially more discriminative information for wildfire segmentation than RGB imagery alone. Across all evaluated architectures, RGB-only configurations produced significantly lower segmentation performance, while infrared-only inputs consistently achieved strong results. These findings highlight the importance of thermal information for active-fire delineation in UAV imagery.

The results also revealed that the effectiveness of multimodal fusion depends on both the selected fusion strategy and the segmentation architecture. Early fusion did not consistently improve performance over infrared-only baselines, while SegFormer with middle fusion achieved the best overall performance. These findings suggest that transformer-based architectures can better exploit RGB--IR information when modality-specific features are learned before fusion. 

Overall, this work provides a systematic analysis of RGB--IR fusion for wildfire segmentation under a unified experimental protocol and establishes strong baseline results for FLAME~3. A limitation of this study is that the evaluation is restricted to the FLAME~3 dataset, and the generalization of the observed trends should be further assessed on additional UAV wildfire datasets acquired under different environmental and sensor conditions.

Future work may investigate additional fusion mechanisms, uncertainty-aware segmentation, radiometric temperature integration, and lightweight multimodal architectures suitable for real-time UAV deployment.


\bibliographystyle{IEEEtran}
\balance
\bibliography{ref.bib}
\balance

\end{document}